\documentclass[conference]{IEEEtran}
\IEEEoverridecommandlockouts
\usepackage{cite}
\usepackage{amsmath,amssymb,amsfonts}
\usepackage{algorithmic}
\usepackage{graphicx}
\usepackage{textcomp}
\usepackage{xcolor}
\usepackage{bm}
\def\BibTeX{{\rm B\kern-.05em{\sc i\kern-.025em b}\kern-.08em
    T\kern-.1667em\lower.7ex\hbox{E}\kern-.125emX}}
\begin{document}

\title{Error-Feedback Model for Output Correction in Bilateral Control-Based Imitation Learning\\
% {\footnotesize \textsuperscript{*}Note: Sub-titles are not captured in Xplore and should not be used}
\thanks{This work was supported by JSPS KAKENHI Grant Number
24K00905, JST, PRESTO Grant Number JPMJPR24T3 Japan and JST ALCA-Next Japan, Grant Number JPMJAN24F1.
This study was based on the results obtained from
the JPNP20004 project subsidized by the New Energy and
Industrial Technology Development Organization (NEDO).}
}

\author{\IEEEauthorblockN{1\textsuperscript{st} Hiroshi Sato}
\IEEEauthorblockA{\textit{Intelligent and Mechanical Interaction Systems} \\
\textit{University of Tsukuba}\\
Tsukuba, Japan \\
sato.hiroshi.tkb\_cu@u.tsukuba.ac.jp}
\and
\IEEEauthorblockN{2\textsuperscript{th} Masashi Konosu}
\IEEEauthorblockA{\textit{Intelligent and Mechanical Interaction Systems} \\
\textit{University of Tsukuba}\\
Tsukuba, Japan \\
konosu.masashi.qa@alumni.tsukuba.ac.jp}
\and
\IEEEauthorblockN{3\textsuperscript{nd} Sho Sakaino}
\IEEEauthorblockA{\textit{Systems and Information Engineering} \\
\textit{University of Tsukuba}\\
Tsukuba, Japan \\
sakaino@iit.tsukuba.ac.jp}
\and
\IEEEauthorblockN{4\textsuperscript{rd} Toshiaki Tsuji}
\IEEEauthorblockA{\textit{Science and Engineering} \\
\textit{Saitama University}\\
Saitama, Japan \\
tsuji@ees.saitama-u.ac.jp}
% \and
% \textit{University of Tsukuba}\\
% Tsukuba, Japan \\
% \and
% \textit{University of Tsukuba}\\
% Tsukuba, Japan \\
}

\maketitle

\begin{abstract}

In recent years, imitation learning using neural networks has enabled robots to perform flexible tasks. However, since neural networks operate in a feedforward structure, they do not possess a mechanism to compensate for output errors. To address this limitation, we developed a feedback mechanism to correct these errors. By employing a hierarchical structure for neural networks comprising lower and upper layers, the lower layer was controlled to follow the upper layer. Additionally, using a multi-layer perceptron in the lower layer, which lacks an internal state, enhanced the error feedback.
In the character-writing task, this model demonstrated improved accuracy in writing previously untrained characters.
% As a result of the character-writing task, an improvement in task accuracy for previously unlearned characters was observed. 
In the character-writing task, this model demonstrated improved accuracy in writing previously untrained characters. Through autonomous control with error feedback, we confirmed that the lower layer could effectively track the output of the upper layer. This study represents a promising step toward integrating neural networks with control theories.

\end{abstract}


\section{Introduction}
% In recent years, with the increasing severity of worker shortages, robots are expected to take over simple tasks in factories.
% However, most current industrial robots are limited to performing repetitive, pre-programmed actions, making it difficult for them to operate in changing environments.
% Therefore, humans are still needed for simple tasks that involve changing environments.

% Recent research has focused on generating robot motion policies using machine learning.
% One example is using reinforcement learning.
% In this method, a reward function is set, and the agent learns it, enabling the robot to grasp various objects\cite{s23073762}\cite{7989385}

% However, reinforcement learning requires designing appropriate reward functions and extensive trial-and-error search, leading to low sample efficiency.

% In contrast, 
% Imitation learning is another approach that learns from human demonstrations\cite{pmlr-v205-zhu23a}\cite{fu2024mobile}
% \cite{ogataken_belcon_IEEE}.
% Imitation learning is a type of supervised learning in which the robot learns from its own previously acquired motion data as the training data.
% This allows the robot to efficiently learn behaviors within the space where the task designer has operated, enabling it to learn complex behaviors with fewer trials than reinforcement learning.

\begin{figure}[!t]
        \centering
        \includegraphics[width=0.85\linewidth]{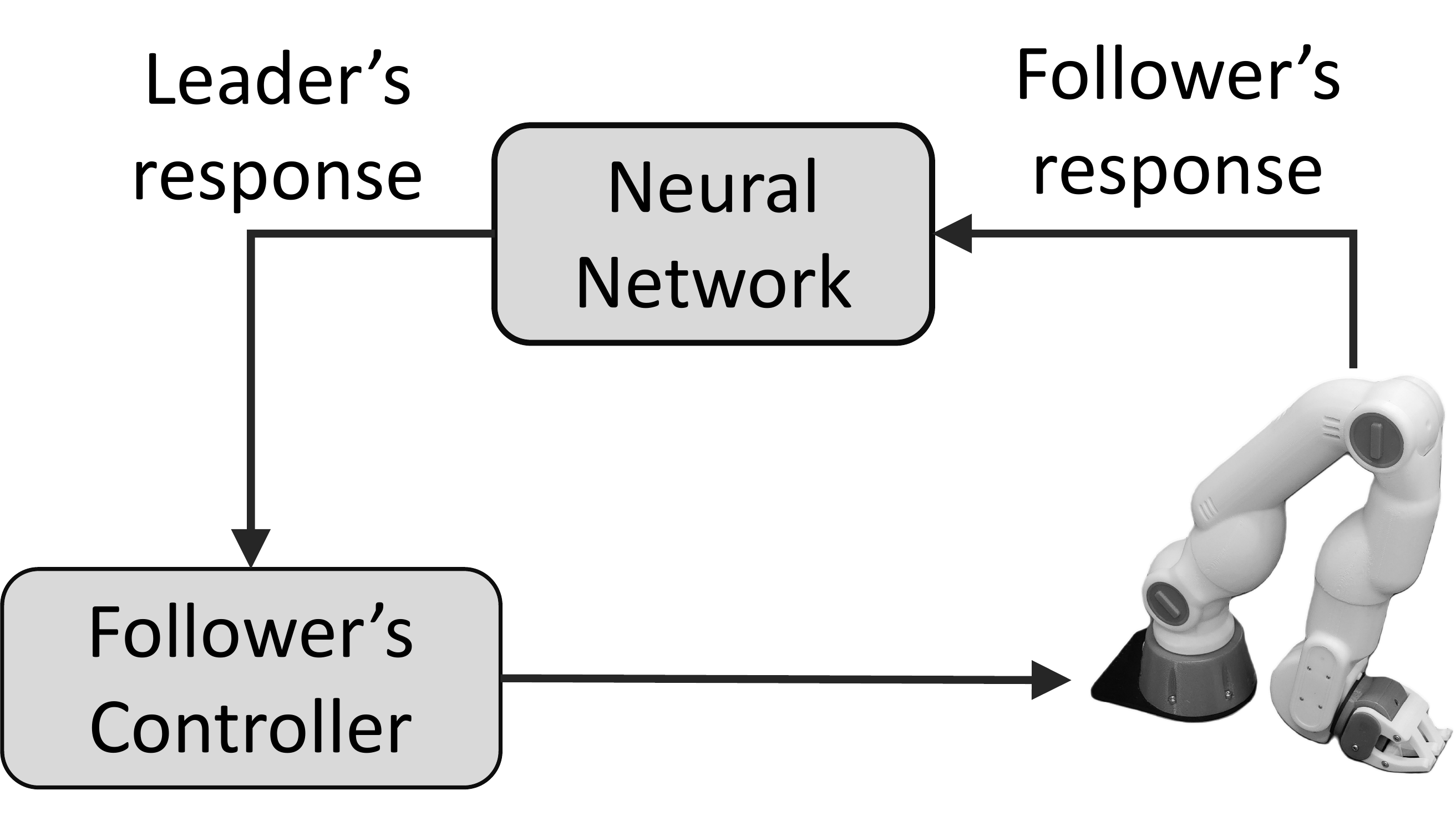}
        \caption{Overview of the autonomous motion using NN}
        \label{fig:autonomous}
\end{figure}

In recent years, imitation learning has gained significant attention for enabling robots to perform complex actions
\cite{pmlr-v205-zhu23a}
\cite{fu2024mobile}
% \cite{ogataken_belcon_IEEE}
\cite{zhao2023learningfinegrainedbimanualmanipulation}.
Imitation learning is a type of supervised learning in which neural networks (NNs) learn from human demonstrations.
% Imitation learning is a type of supervised learning in which the robot learns from previously acquired motion data performed by a human as the training data. 
% This allows the robot to learn behaviors in the space where the task designer has manipulated the robot, enabling it to learn complex behaviors with fewer trials.
Furthermore, research on imitation learning using position and force information has advanced.
% In recent years, research on imitation learning using both position and force information has advanced.
 Specifically, bilateral control-based imitation learning has proven effective in reproducing human force application
% \cite{adachi2018imitation}
\cite{saigusa_nonprehensile}
\cite{AKAGAWA_HPF}
\cite{yamaneHand}
\cite{buamanee2024biactbilateralcontrolbasedimitation}.
Bilateral control is a teleoperation technology that uses two robots: one interacts with the environment, while the other is operated by a human applying force. By collecting data with this technology, both position and force response and command values can be obtained, allowing robots to replicate human operational sensations. The use of force-based imitation learning shows promise for replacing human tasks with robots. 
% In this study, we focus on imitation learning using bilateral control. 
% For details on the method, refer to Chapter 2.
However, in conventional bilateral control-based imitation learning, NN has a feedforward structure and does not control output errors, as shown in Fig.~\ref{fig:autonomous}. 
Hence, errors during the autonomous operation of the NN are not compensated. This issue is observed not only in bilateral control-based imitation learning but also in many NN-based imitation learning approaches.

Traditionally, NNs have required an internal state to retain memory for handling time-series data. However, this NN struggles to integrate with controllers due to the significant influence of the internal state. This suggests that the system's non-Markovian nature complicates NN control. Generally, systems are more likely to exhibit Markovian properties when the sampling period is shortened. Therefore, to realize effective NN control, it is essential to establish a structure with independent components for the short-sampling-period NN, which exhibits Markovian properties and is easier to control, and the long-sampling-period NN, which has non-Markovian properties and enables complex time-series inference.

In this study, we developed a control system for a hierarchical model with different sampling periods.
The hierarchical model comprises an upper layer that makes long-term predictions and a lower layer that makes short-term predictions.
The upper layer is a strong non-Markovian system that predicts action plans based on past information.
Conversely, the lower-layer is a strong Markovian system that predicts command values and states with a short sampling period. 
This type of model has been proposed in previous research\cite{hayashi2022independently}, demonstrating its effectiveness for long-term tasks. 
However, prior studies employed Long Short-Term Memories (LSTMs) with internal states for the Markovian lower-layer. 
Therefore, we employed a multilayer perceptron (MLP), which lacks an internal state, to construct a control system for the output. 
During the control process, the error in the robot’s state predicted by the upper and lower layers was fed back into the input of the lower layer. This allows the lower layer to adjust its inference to minimize the error relative to the upper layer's predictions. We refer to this model as the error feedback model.

The effectiveness of the proposed method was validated through a character-writing task involving both learned and unlearned characters. Evaluation was based on the accuracy of the drawn characters and the trajectory of angles. For comparison, the lower layer was implemented using both LSTM and MLP. It is important to note that this study explores the potential of the error-feedback model for the lower layer, assuming that the state predicted by the upper layer is already known. This approach is expected to lead to the development of a new model that combines NNs with control systems.

\section{RELATED WORKS}

% \subsection{NewtonianVAE}
% NewtonianVAE is a method that combines control with NNs (NNs). It maps high-dimensional data, such as images, to a latent space in NNs, imposing constraints based on state equations. This approach enables proportional control in robot operations relying on visual information, significantly simplifying vision-based imitation learning\cite{9577560}\cite{9981610}.
% By using NewtonianVAE, Miguel et~al. demonstrated proportional control feasibility in the latent space for a 'G' shape drawing task, while Okumura et~al. successfully corrected positional errors in the latent space to improve insertion success rates in a USB plug-insertion task.
% However, NewtonianVAE places dynamic assumptions based on Newtonian motion equations in the latent space, assuming consistent target dynamics. Consequently, tasks involving complex interactions with objects or a combination of multiple distinct movements, where actions vary over time, are challenging to represent in the latent space. In such cases, applying NewtonianVAE is difficult, and task performance itself may be hindered.

\subsection{World Model with Control System}
Integrating control with NNs has been extensively studied using world models\cite{NIPS2015_a1afc58c}\cite{ha2018worldmodels}\cite{9577560}. A world model is an NN that learns the structure of the environment from observation data, representing it in a latent space. By incorporating a mechanism to control errors in this latent space, it becomes possible to combine NNs with control. However, these methods assume constant dynamics, which poses challenges for tasks involving contact or multiple actions, as dynamic changes over time complicate mapping to the latent space.

\subsection{Bilateral Control-Based Imitation Learning}
Bilateral control-based imitation learning uses bilateral control during the data collection phase. Bilateral control is a teleoperation technique that synchronizes the positions and forces of two robots: a leader and a follower. The leader receives forces from a human operator, while the follower interacts with the environment. Using this control method to perform tasks, the response values of both the leader and follower are collected. The leader’s response value serves as the command for the follower. As a result, the response and command values of the follower can be collected separately. An NN is then trained to predict the next command value of the follower based on its current response value. Once trained, the NN enables autonomous movements that replicate bilateral control, allowing the execution of tasks requiring force control.

\subsection{Hierarchical Model}
\begin{figure}[!t]
    \centering
        \centering
        \includegraphics[width=0.9\linewidth]{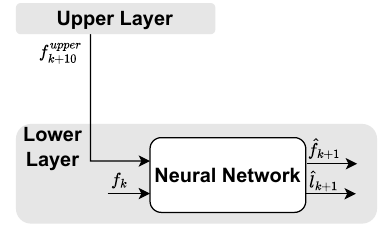}
        \caption{Hierarchical model proposed by Hayashi $et~al.$\cite{hayashi2022independently}}
        \label{fig:ordinary_model}
\end{figure}

A hierarchical model processes information at different levels of abstraction in each-layer, breaking down complex tasks into manageable sub-tasks.
\cite{10295965}
\cite{hayashi2022independently}.
Hayashi $et~al.$ proposed a hierarchical model for bilateral control-based imitation learning
\cite{hayashi2022independently}.
The proposed hierarchical model is shown in Fig.~\ref{fig:ordinary_model}.
Here, $\bm{f}$ and $\bm{l}$ represent the follower and leader, respectively, and the subscript $t$ indicates the operational step of the NN.
The upper-layer infers the state 10 steps ahead $\bm{f}^{\text{upper}}_{k+10}$ and provides it to the lower-layer.
Conversely, the lower-layer considers the follower's current state $\bm{f}_k$  
and state $\bm{f}^{\text{upper}}_{k+10}$ provided by the upper-layer as inputs.  
It then predicts the follower's next state $\bm{\hat{f}}_{k+1}$ and leader's next state $\bm{\hat{l}}_{k+1}$.

The hierarchical model has been shown to be effective for long-term tasks. Additionally, it has been confirmed that the model can accomplish tasks even when the lower layer receives unlearned information from the upper layer. However, without a control mechanism for the lower layer to follow the upper layer, there is concern about a decrease in task performance accuracy.

\section{Proposed Method}

% \begin{figure}[!t]
%         \centering
%         \includegraphics[width=\linewidth]{figures/position_force_control.pdf}
%         \caption{Block diagram of robot control system}
%         \label{fig:control_system}
% \end{figure}
\subsection{Separation of the Markovian and non-Markovian properties of tasks}
In this study, a hierarchical model was employed to separate the Markovian and non-Markovian aspects of the system.
The hierarchical model is the same as those used in previous research\cite{hayashi2022independently}.
The upper-layer infers task plans over a long period, handling the non-Markovian properties of the system.  
In contrast, the lower-layer performs short-period inference of commands and states  
based on the current state and a few steps ahead of the follower's state.
Given that it performs short-period inference and involves predictions that interpolate between different steps,  
the lower-layer system is considered to exhibit high Markovian properties.
Based on this reasoning, control was constructed for the Markovian aspects of this hierarchical model.

\begin{figure}[!t]
        \centering
        \includegraphics[width=0.9\linewidth]{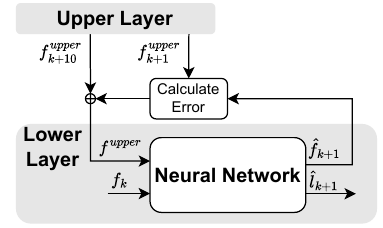}
        \caption{Proposed error-feedback model}
        \label{fig:Proposed_NN_Model}
\end{figure}

\subsection{hierarchical model with error-feedback mechanism}
% The purpose of this study is to incorporate control into an NN.  
In this study, we propose an error feedback model, as shown in Fig.~\ref{fig:Proposed_NN_Model}.
This model is designed to control the system to reduce the error between the outputs.
\color{black}
In the proposed method, the upper-layer generates the state one step ahead \(\bm{f}^{\text{upper}}_{k+1}\) and state ten steps ahead \(\bm{f}^{\text{upper}}_{k+10}\).
Additionally, the error between the output of the lower-layer \(\hat{\bm{f}}_{k+1}\) and upper-layer output \(\bm{f}^{\text{upper}}_{k+1}\) was calculated. 
Here, the error calculation is defined simply as the difference \(\bm{f}^{\text{upper}}_{k+1} - \hat{\bm{f}}_{k+1}\). Extending this to various control mechanisms is a future work, but the simple difference worked well in this study.
Subsequently, the information from the upper-layer given to the lower-layer, \(\bm{f}^{\text{upper}}\), is defined as follows:
\begin{equation}
\label{eq:A.E.}
\bm{f}^{\text{upper}} = \bm{f}^{\text{upper}}_{k+10} + \left( \bm{f}^{\text{upper}}_{k+1} - \hat{\bm{f}}_{k+1} \right)
\end{equation}  
By adding the error to the upper-layer output, it is expected that the lower-layer will generate an output that has been corrected for errors.

It should be noted that \(\bm{f}^{\text{upper}}_{k+10}\) is updated every ten steps, while \(\bm{f}^{\text{upper}}_{k+1}\) is updated at each time step.  
This was standardized to enable comparison with previous research\cite{hayashi2022independently}.
% Although \(\bm{f}^{\text{upper}}_{k+10}\) could also be updated at each time step,  
% it was aligned with previous research for comparison purposes \cite{hayashi2022independently}.  
Additionally, this mechanism was applied during the robot's autonomous operation. Therefore, the learning process does not include an error feedback mechanism, similar to conventional hierarchical models.

\subsection{NN Design of the lower-layer}
\begin{figure}[!t]
    % \begin{minipage}[t]{0.38\linewidth}
        \centering
        \includegraphics[width=0.8\linewidth]{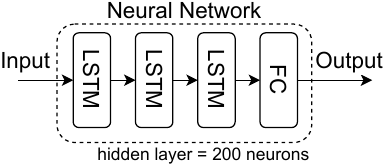} % 高さを指定
        \caption{LSTM model}
        \label{fig:LSTM_model}
    % \end{minipage}
\end{figure}

\begin{figure}[!t]
    % \hfill % 空白を入れて横に並べる
    % \begin{minipage}[t]{0.58\linewidth}
        \centering
        \includegraphics[width=0.8\linewidth]{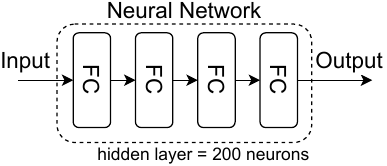} % 高さを指定
        \caption{MLP model}
        \label{fig:MLP_model}
    % \end{minipage}
\end{figure}

To address the Markovian properties of the system, the lower-layer is designed as a simple multi-layer perceptron (MLP) without internal states, as illustrated in Fig.~\ref{fig:MLP_model}.
The network comprises four layers of fully connected layers with 200 dimensions using the Tanh function as the activation function for all layers except for the final one.  
Additionally, for comparison, the lower layer using the conventional long short-term memory (LSTM) architecture is also presented in Fig.~\ref{fig:LSTM_model}.  
The LSTM network is comprised of three layers of 200-dimensional LSTM units and a fully connected layer, resulting in a total of four layers.

In this study, the states predicted by the upper-layer are assumed to be known.
% This is done to ensure that the same information is provided from the upper-layer to the lower-layer for each autonomous operation.  
% By providing the same information, the reproducibility of the experiments is enhanced, allowing for a performance comparison of error feedback in the lower-layer.  
Specifically, time-series states of the follower are stored in advance through bilateral control, and this data is utilized.  
This allows for the comparison of different lower-layers using the same predictions from the upper-layer.
% It is expected that using states generated by a model based on NNs in the upper-layer will improve the generalizability of the task; however, the verification of this remains a future challenge.


\section{Experiment Method}
\subsection{Manipulator}
In this study, CRANE-X7, manufactured by RT Corporation, was employed.
The manipulator has seven degrees of freedom, and the gripper has one degree of freedom.
The gripper was replaced by a cross-structured hand\cite{yamaneHand}.
Given that controlling a manipulator with seven degrees of freedom can be challenging for humans, joint 2 was fixed using position control, effectively reducing the system to a six degrees of freedom manipulator. 
Each axis of the manipulator was controlled by a position and force hybrid controller, with a control period of 500 Hz
\cite{saigusa_nonprehensile}.
% Since the 7 degrees of freedom manipulator is difficult to control by human, joint 2 was fixed by position control and used as a 6 degrees of freedom manipulator.
% Each axis of the manipulator was controlled by a
% position and force hybrid controller.
% The control period was 500~Hz. 
The joint angles $\bm{\theta}$ were measured by rotary encoders at each joint, and the angular velocities $\Dot{\bm{\theta}}$ were calculated by its pseudodifferential.
The torques $\bm{\tau}$ were estimated using a reaction force estimation observer
\cite{murakami1993torque}.
% as shown in Figure \ref{fig:control_system}.
% In this figure, $\bm{\theta}$, $\Dot{\bm{\theta}}$, and $\bm{\tau}$ represent the angles, angular velocities, and torques at each joint of the robot.
% Also, superscripts $cmd$, $ref$, $res$, and $dis$ indicate command value, reference value, response value, and disturbance.
% The disturbance torque $\bm{\tau}^{dis}$ was calculated by the disturbance observer (DOB)\cite{ohnishi1996motion}. The torque response $\bm{\tau}^{res}$ was estimated using a reaction force estimation observer (RFOB)\cite{murakami1993torque}.

\subsection{Verification of autonomous operation}
In the experiment, a writing task was conducted using the robot.
The robot wrote the characters while holding the pen from the beginning, as shown in Fig.~\ref{fig:task_overview}.
% The pen was attached to a jig made by a 3D printer, as shown in Fig.~\ref{fig:jig}, to prevent it from being displaced by the robot.
The characters drawn by the robot were captured by an Intel RealSense D435i mounted on the top of the whiteboard.

\subsubsection{Preliminary}
As a preliminary experiment, we investigated the amount of information required as the upper-layer output.
Specifically, in Fig. 3, the upper layer outputs one of the following three:
% Specifically, the upper-layer outputs in the figure were defined as
$\bm{f}_{k+10}^{upper}=[\bm{\theta}],~\bm{f}_{k+10}^{upper}=[\bm{\theta}, \Dot{\bm{\theta}}], ~\bm{f}_{k+10}^{upper}=[\bm{\theta}, \Dot{\bm{\theta}}, \bm{\tau }]$.
The lower-layers were trained to write the character 'A' using two model types: LSTM and MLP.
\color{black}
% Since the purpose of this study is to control the lower-layers, the output $\bm{f}$ of the upper-layers were the follower response values collected in advance by the bilateral control.
With the learned NNs, the robot performed the operation of writing 'A.'
In the autonomous operation, the upper-layer outputs, which were collected in advance by the bilateral control, were used.
The optimal amount of information for the upper-layer outputs were selected by the evaluation described below.

\subsubsection{Evaluation of error-feedback model}
A comparison was conducted between the conventional hierarchical model and proposed hierarchical model.
The lower-layer models, which were highly evaluated in the preliminary experiments, were utilized. 
An experiment was conducted to write three different types of characters: character 'A,' '4,' and 'B.' 
Specifically, 'A' is a learned character, while '4' and 'B' are characters that had not been learned.

Character '4' was selected because it has a shape that is similar to 'A' but somewhat different, while character 'B' was chosen for its distinct shape compared to 'A.'
To enable the writing of these characters, the upper-layer outputs were modified to correspond to each respective character.
The upper-layer outputs was determined based on the follower’s state, which was previously collected through bilateral control.

Meanwhile, the lower-layer NN remained unchanged from the preliminary experiments. In conventional imitation learning, the NN must be retrained for each new character. When the upper-layer outputs for an unlearned character are used, errors are expected in the NN’s output. Therefore, we aimed to verify whether the proposed method could suppress these errors and generate command values that align with the upper-layer outputs.

% The character 4 was selected because it has a similar shape to A but is somewhat different, and B was chosen for its distinct shape compared to A. 
% To enable the writing of these characters, the upper-layer outputs was modified to correspond to the respective characters, while the lower-layer NNs were kept unchanged from the preliminary experiments.
% The outputs of the upper-layer for 4 and B were the follower’s response values collected through bilateral control in advance.

% Since 4 and B are unlearned characters, the output of the NNs is expected to have errors. In conventional imitation learning, the NNs need to be retrained for each character. Therefore, we aimed to verify whether the proposed method can suppress these errors and generate command values that follow the upper-layer outputs.

% 4 was selected because it has a similar shape to A but is somewhat different, and B was selected for its different shape from A.

% To enable the writing of these characters, the lower-layer NN were kept unchanged from the preliminary experiments, while the upper-layer outputs was changed to the respective characters.
% The outputs of the upper-layer for “4” and “B” were the follower’s response values collected by the bilateral control in advance.

% Since “4” and “B” are unlearned characters, the output of the NN is expected to have errors.
% So conventional imitation learning need to re train NN for each characters.
% Therefore, we verified whether the proposed method can suppress the errors and generate command values that follow the upper-layer.

\begin{figure}[!t]
    \centering
    \begin{minipage}{0.55\linewidth}
        \centering
        \includegraphics[height=4.2cm]{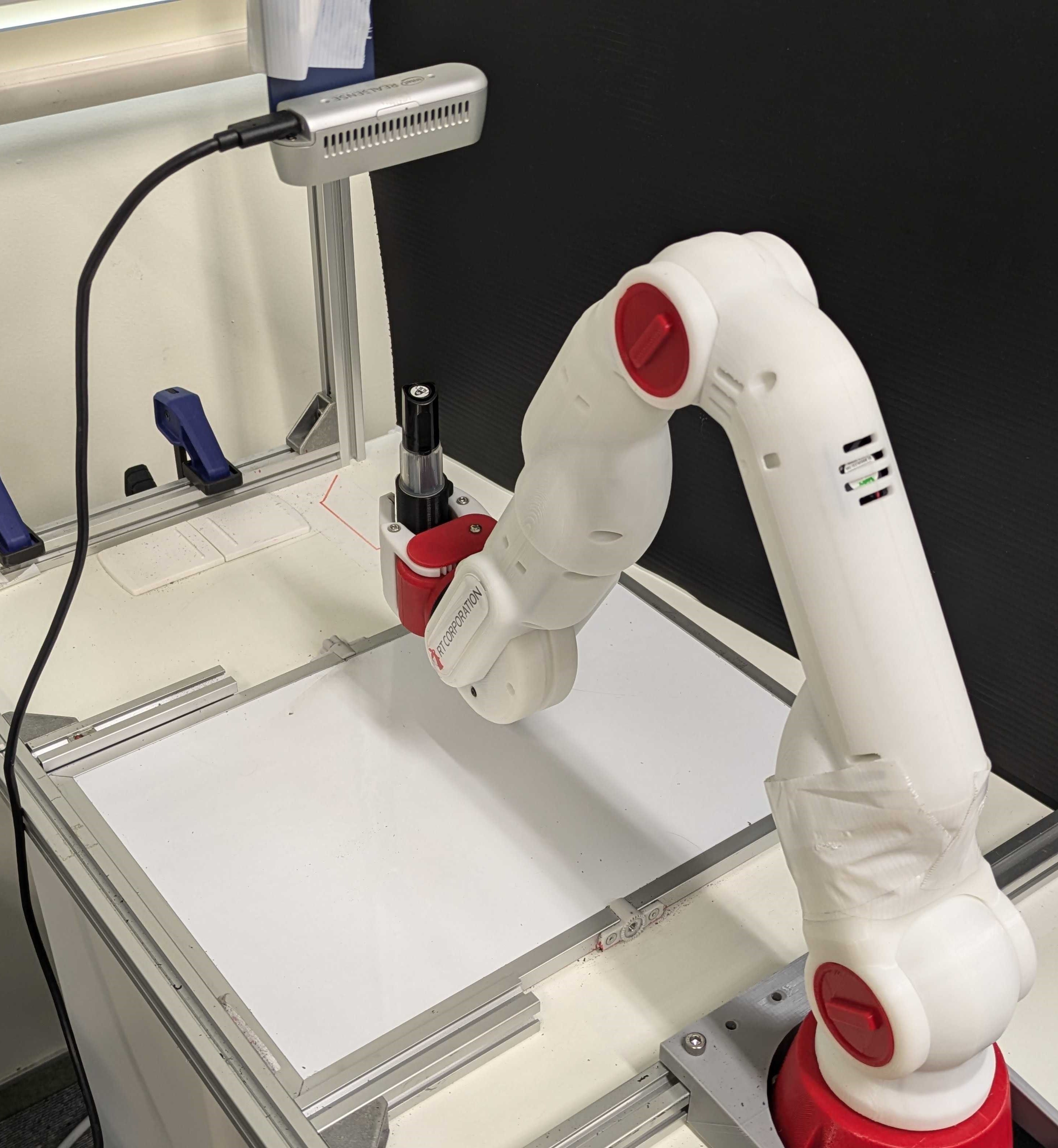} 
        \caption{Task environment}
        \label{fig:task_overview}
    \end{minipage}
    % \begin{minipage}{0.1\linewidth}
    %     \centering
    %     \includegraphics[width=\linewidth]{figures/jig.pdf} 
    %     \caption{jig}
    %     \label{fig:jig}
    % \end{minipage}
    \begin{minipage}{0.43\linewidth}
        \centering
        \includegraphics[height=4.2cm]{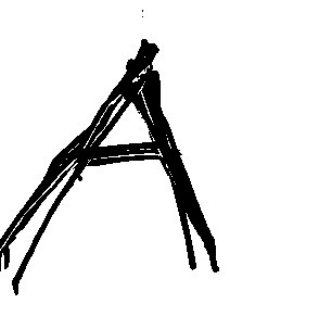} 
        \caption{Characters drawn by training data}
        \label{fig:training_data}
    \end{minipage}
\end{figure}

\subsection{Training NN}
% The NNs used in this study were Fig.~\ref{fig:LSTM_model} and Fig.~\ref{fig:MLP_model}. 
% The LSTM consisted of a 3-layer 200-dimensional LSTM and fully-connected (FC)-layers as shown in Fig.~\ref{fig:LSTM_model}.
% The MLP consisted of 4-layer 200-dimensional FC-layers as shown in Fig.~\ref{fig:MLP_model}.
% The hyperbolic tangent function (Tanh) was used as the activation function except for the final-layer.
We collected training data and validation data by using bilateral control to write character 'A.' 
The training data were collected seven times, five of which were used as training data and two as validation data.
Fig.~\ref{fig:training_data} is a diagram that is drawn when the data is collected.
This is the image of the characters on the whiteboard captured by a camera and superimposed after binarization.

When training NN using time-series data, the learning efficiency can be improved by reducing the sampling frequency. For this purpose, the joint information acquired at 2-ms intervals was sampled every 10 steps by shifting the starting point, creating a data set with 20-ms intervals\cite{rahmatizadeh2018virtual}.
This process increased the amount of data by a factor of 10.

Additionally, a 20~rad/s low-pass filter was applied to the teacher data to remove high-frequency components.
The input data were augmented with normally distributed noise with a variance of 0.01.
The length of each sequence was unified by padding, which copies the last value.
The data were normalized to mean 0 and standard deviation 1 during training.
Mean Squared Error (MSE) was used as the loss function and Adam was used for optimization. The learning rate was set to 0.0001, and the batch size to 16, and the number of epochs to 1000.

\subsection{Evaluation Method}
Evaluation of autonomous movements was performed in two ways: assessing the diagrams drawn by the robot and evaluating the joint angles during autonomous movements. 
Autonomous movements were performed five times, and the means and standard deviations were calculated for both evaluation methods. 
The outputs from the upper-layer in this study were collected using bilateral control. Therefore, the upper-layer outputs of the drawn characters and joint angle information were collected in advance.

\subsubsection{IoU} 
Intersection over Union (IoU) was used to evaluate the accuracy of the diagrams.
The diagrams drawn by the robot were captured by a camera and binarized in black and white.
The IoU is calculated as follows:
\begin{equation}
\label{eq:IoU}
IoU = \frac{B^{upper} \cap B^{output}}{B^{upper} \cup B^{output}}
\end{equation}
where $B^{upper}$ represents the character area drawn in the upper-layer output, and $B^{output}$ denotes the character area drawn by the autonomous motion of NN.
As this value approaches 1, it indicates a higher degree of match between the two characters.
By comparing the IoU values, we evaluated whether the autonomous control followed the upper-layer output.

\subsubsection{MSE of Angles}
The joint angles of autonomous movement were evaluated.
The purpose of this study is to control the lower-layer to approach the upper-layer outputs.
In other words, by calculating the error between the joint angles obtained by autonomous movements and joint angles of the upper-layer output, it is possible to evaluate the follow-up to the upper-layer outputs.
We termed this as Angular Error and calculated it as follows:

\begin{equation}
    \label{eq:Angle_error}
    Angular Error = 
    \sum_{k=0}^{n} \left( \bm{{\theta}}^{upper}_k - \bm{{\theta}}^{res}_k \right)
\end{equation}
where $k$, $k=0$, $k=n$, $\bm{{\theta}}^{upper}$, and $\bm{{\theta}}^{res}$ denote specific time, task start time, task end time, the angle of the upper-layer output, and the angle response value of the robot during autonomous operation.
By comparing these values, we evaluated the tracking performance of the robot relative to the upper-layer output.

\begin{figure}[!t]
        \centering
        \includegraphics[width=0.95\linewidth]{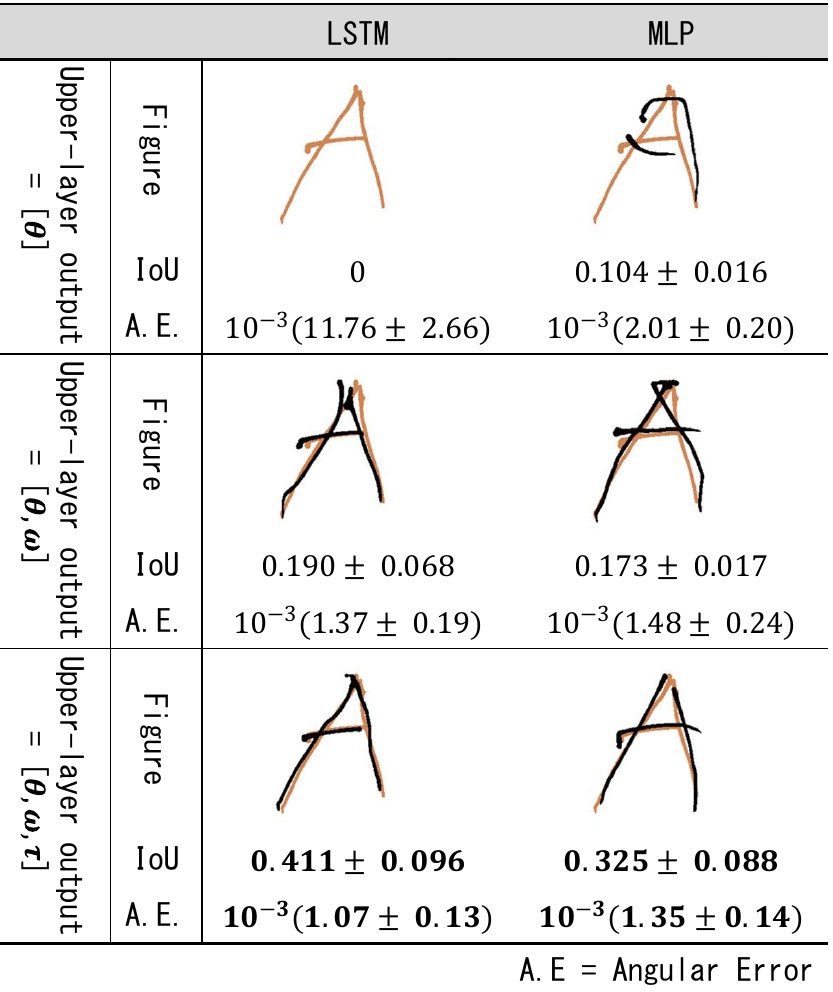}
        \caption{Performance comparison of autonomous robot operation based on differences in upper-layer information}
        \label{fig:pre_results}
\end{figure}
\begin{figure*}[!t]
    \centering
    \includegraphics[width=0.90\linewidth]{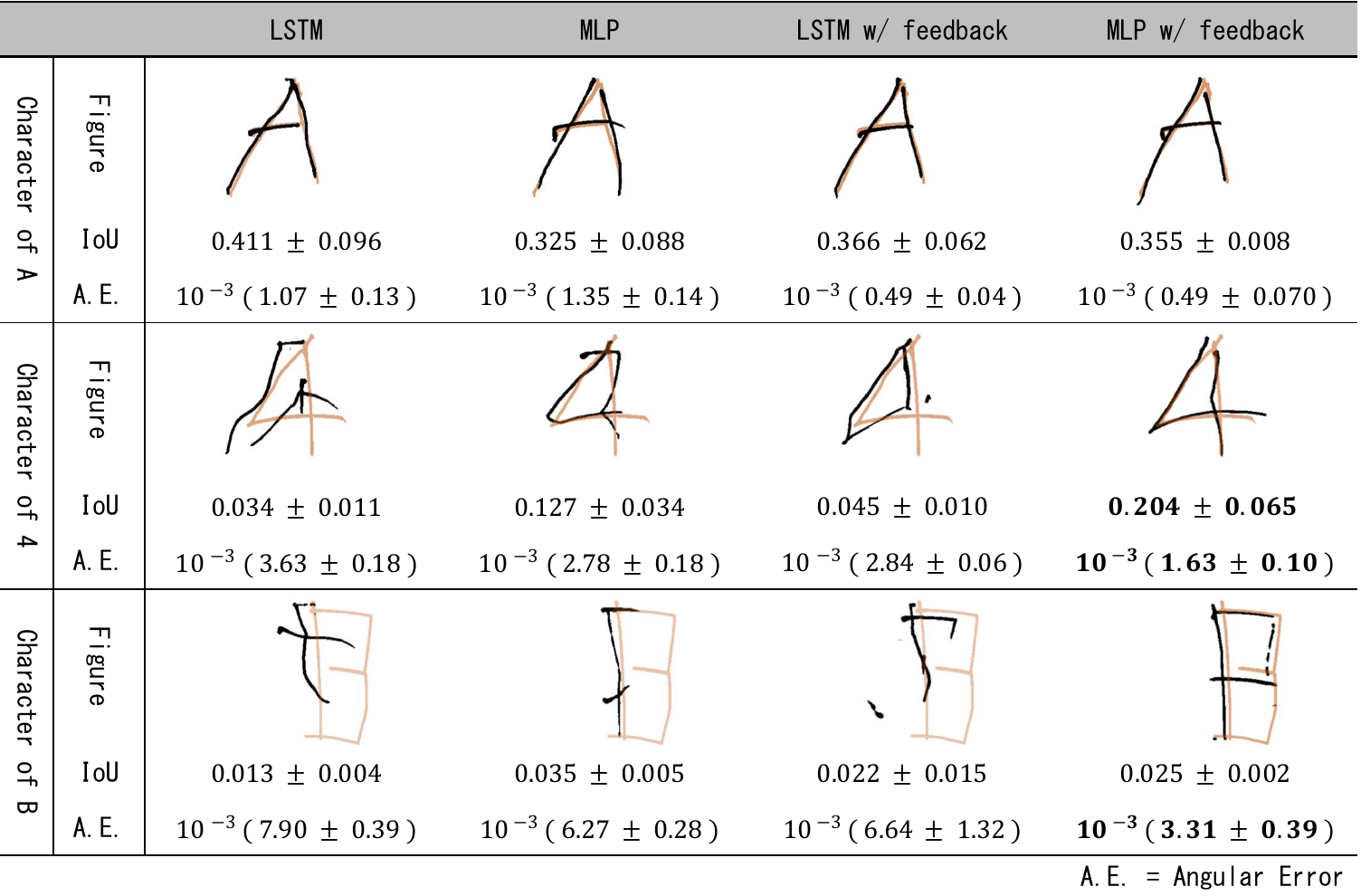}
    \caption{Comparison of Autonomous Performance: error-feedback model in Writing 'A,' '4,' and 'B'}
    \label{fig:results}
\end{figure*}

% 表と図をtexで作りたかったところである
% \begin{table}[!t]
%     \caption{Performance Comparison of Autonomous Robot Operation Based on Differences in Information Quantity from upper-layers}
%     \begin{center}
%     \begin{tabular}{c|c|c|c}
%         \hline
%         & & \textbf{\textit{LSTM}}& \textbf{\textit{MLP}} \\
%         \hline
%         \rotatebox{-90}{Upper = {$[\theta]$}} & \rotatebox{-90}{figure} & \includegraphics[width=0.3\linewidth]{figures/B_mlp.png} & \includegraphics[width=0.3\linewidth]{figures/B_mlp.png} \\
        
%         % \hline
%         \cline{2-4}
%         & IoU & &  \\
%         \cline{2-4}
%         % \hline
%         & A.E.$^{\mathrm{*}}$ & &  \\
%         \hline
        
%         \multicolumn{4}{l}{$^{\mathrm{*}}$A.E = Angular Error}
%         \end{tabular}
%         \label{tab1}
%     \end{center}
% \end{table}

\section{EXPERIMENT}
\subsection{Preliminary}
The results of the preliminary experiments are shown in Fig.~\ref{fig:pre_results}.
The black line in the figure represents the drawing made by autonomous actions, while the light red color indicates the upper-layer output.
% IoU and Angular error are the mean and standard deviation of the five trials of the robot.
Comparing the drawn characters with the IoU, LSTM and MLP achieved the highest values when using $\bm{f}_{k+10}^{upper} = [\bm{\theta}, \Dot{\bm{\theta}}, \bm{\tau}]$ as the upper-layer outputs. Additionally, the angle error was lowest under the same conditions.

It is considered that this task required future force because writing characters involves applying force to a board. 
% Therefore, it is believed that the upper-layer output, which included torque, achieved the highest score.
Based on these results, $\bm{f}_{k+10}^{upper} = [\bm{\theta}, \Dot{\bm{\theta}}, \bm{\tau}]$ was used as the upper-layer output to verify the effectiveness of the error-feedback model.

\subsection{Evaluation of error-feedback model}
To verify the effectiveness of the proposed method, tasks were conducted to write several characters: 'A,' '4' and 'B.'
The results with the error-feedback model are described as 'w/ feedback.'
% The upper-layer outputs were set to $\bm{f}_{k+10}^{upper}=[\bm{\theta}, \Dot{\bm{\theta}}, \bm{\tau}]$ which was the most effective model from the preliminary experiments. The results of the autonomous actions are shown in Fig.~\ref{fig:results}.

\subsubsection{character A}

The upper part of Fig.~\ref{fig:results} presents the results for the task of writing character 'A.' Using the error-feedback model, no increase in IoU was observed for either LSTM or MLP, though a decrease in Angle Error was confirmed. 
% While there was no significant difference between LSTM and MLP, the LSTM model performed slightly better. 
In the error-feedback model, there was no significant difference between LSTM and MLP, but the LSTM model showed slightly higher IoU results.
Overall, it can be concluded that the conventional LSTM provides sufficient performance in writing character 'A,' as the lower-layer have learned these characters. 
The observed decrease in Angle Error, while not affecting the IoU, suggests successful approximation of the upper-layer output.
\color{black}
% The results for the task of writing the character 'A' are presented in the upper part of the figure.
% Through autonomous motion using the error-feedback model, no increase in IoU was observed for either LSTM or MLP, while a decrease in Angle Error was confirmed.
% When comparing LSTM and MLP, no significant difference was observed between them, but the results of LSTM were slightly better.

% It can be concluded that using the conventional LSTM provides sufficient performance.
% This is because the lower layer have learned the characters.
% However, a decrease in Angle Error was observed in both LSTM and MLP. Although this did not affect the results of the A-writing task, it shows that the objective of approximating the upper-layer output has been achieved.
\subsubsection{character 4}
The middle part of the Fig.~\ref{fig:results} presents the results for the task of writing character '4.'
In the error-feedback model, MLP showed an increase in IoU, while LSTM showed a slight increase. 
Similarly, MLP reduced the Angle Error, and LSTM showed a slight reduction. 
When comparing the performance of LSTM and MLP, MLP achieved a larger IoU and a smaller Angle Error than LSTM.
These results indicate that using MLP with the error-feedback model improved task accuracy. 
Given that MLP does not have an internal state, it made appropriate predictions based on error feedback without being influenced by past memories.

\subsubsection{character B}
The bottom part of the Fig.~\ref{fig:results} presents the results for the task of writing character 'B.'
In the error-feedback model, no increase in IoU was observed for either LSTM or MLP. However, a decrease in Angle Error was noted.
When comparing LSTM and MLP, MLP exhibited a smaller Angle Error than LSTM.

These results suggest that NN model faced difficulties in performing the task of writing the character 'B.'
However, with the error-feedback model using MLP, there was a significant reduction in Angle Error.
This indicates that the followability to the upper-layer outputs improved.
The low IoU values are thought to be due to the pen not making contact with the board, despite following the upper-layer outputs.

The difficulty in performing the task of writing the character 'B' may be attributed to the fact that the motion for 'B' involved extrapolation beyond the learned data.
In particular, the second stroke falls outside the range of the motion used for drawing 'A.'
As a result, the lower-layer may not have learned the skills necessary for writing 'B,' making it challenging to follow the upper-layer output.

% In the autonomous operation with feedback, both LSTM and MLP had smaller IoU values and larger Angular Error than A. In the autonomous operation with feedback, there was no significant improvement in IoU values. Also, the autonomous operation with feedback did not significantly improve the IoU values.
% On the other hand, the MLP with feedback resulted in a significant reduction in Angular Error.

% This result indicates that the MLP follows the upper-layer output better than the LSTM in autonomous operation with feedback.
% The reason for the low IoU value seems to be that the trajectory of the pen did not hit the board, even though it followed the the upper-layer output.
% In addition, it is considered that the motion of B was an extrapolated motion of the training data.
% In addition, it is considered that the motion of B was an extrapolated motion of the training data.
% In particular, the second stroke that draws B is mainly out of the range of the stroke A.
% Therefore, it is thought that the lower-layer were not able to learn the skills required to draw B, making it difficult for them to follow the the upper-layer output.
\section{Conclusion}
% \section{Discussion}
% In this study, we proposed an error-feedback model for a hierarchical NN that incorporates a feedback system to address output errors.
% In the proposed method, output errors between the upper and lower-layer were calculated and incorporated into the input of the lower-layer.
% In the writing task experiments, the error-feedback model accurately followed the upper-layer output. 
% Additionally, using MLP in the lower-layer improved the tracking performance relative to the upper-layer output, contributing to improved accuracy in character generation.
% These results show the potential as a first step toward the integration of NN and control theories.
In this study, we proposed an error-feedback model for a hierarchical NN that addresses output errors through feedback. 
The method calculates output errors between upper and lower layers and incorporates them into a lower-layer input. 
In the writing task experiments, the model accurately followed the upper-layer output. 
Additionally, using an MLP in the lower-layer enhanced tracking performance and improved the accuracy of character generation.
It is believed that these results represent a significant first step toward integrating NNs with control theory.

\section{Future Works}
The next step is to develop an lower-layer that can more accurately follow the upper-layer output.
In this study, the lower-layer learned to write only character 'A.'
To enable the lower-layer to more flexibly follow the upper-layer output, it is necessary for lower layer to learn a wider range of behaviors. Thus, it is essential to verify whether teaching a variety of behaviors to the lower-layer improves their ability to follow the upper-layer output.

Additionally, the next challenge is the extension to the lower-layer, where only the trajectory is provided from the upper-layer.
In this study, the outputs of the upper-layer were angle, angular velocity, and torque. 
However, in many cases, it is difficult to obtain all three of these physical quantities. 
If the lower-layer only require the angle information from the upper-layer, it becomes possible to combine this approach with methods such as direct teaching.
To realize this, a lower-layer capable of generating effective outputs from limited information is required. 
For example, by using generative models, such as Conditional Variational Autoencoders (CVAE) \cite{NIPS2014_d523773c}, it is expected that other physical quantities can be generated from angle information.
These development can lead to further advancements in the field of robotics.

% \section*{Acknowledgment}
% This work was supported by JSPS KAKENHI Grant Number 24K00905, JST, PRESTO Grant Number JPMJPR24T3 Japan and JST ALCA-Next Japan, Grant Number JPMJAN24F1.
% This study was based on the results obtained from the JPNP20004 project subsidized by the New Energy and Industrial Technology Development Organization (NEDO).
% \color{black}

\bibliographystyle{IEEEtran}
\bibliography{ICM_sato.bib}

\vspace{12pt}

\end{document}